\documentclass[Afour,sagev,times, doublespace]{sagej}

\usepackage{moreverb,url}
\usepackage{lmodern}

\usepackage[colorlinks,bookmarksopen,bookmarksnumbered,citecolor=red,urlcolor=red]{hyperref}

\newcommand\BibTeX{{\rmfamily B\kern-.05em \textsc{i\kern-.025em b}\kern-.08em
T\kern-.1667em\lower.7ex\hbox{E}\kern-.125emX}}

\usepackage{bm}
\usepackage{comment}

\usepackage{subcaption}
\usepackage{graphicx}

\def\volumeyear{202}

\newcommand{\sss}{\scriptscriptstyle}
\DeclareMathOperator*{\argmax}{arg\,max}
\DeclareMathOperator*{\argmin}{arg\,min}

\makeatletter
\def\@maketitle{%
\if@Royal \vspace*{-20pt} \fi
\if@Crown \vspace*{-20pt} \fi
\vspace*{-34pt}%
\null%
\begin{center}
\if@PCfour \begin{rm} \else \begin{sf} \fi
\begin{minipage}[t]{\textwidth}
  \vskip 12.5pt%
    {\raggedright\titlesize\textbf{\@title} \par}%
    \vskip 1.5em%
    \vskip 12.5mm%
\end{minipage}
{\par\large%
\if@Royal \vspace*{6mm} \fi
\if@Crown \vspace*{6mm} \fi%
\lineskip .5em%
{\raggedright\textbf{\@author}\par}}
\vskip 40pt%
{\noindent\usebox\absbox\par}
{\vspace{20pt}%
{\noindent\normalsize\@keywords}\par}
\if@PCfour \end{rm} \else \end{sf} \fi
\end{center}
\if@Royal \vspace*{-4.5mm} \fi
\if@Crown \vspace*{-4.5mm} \fi
\vspace{22pt} \par%
}

\def\ps@sagepage{%
\let\@mkboth\@gobbletwo
\def\@evenhead{\parbox{\textwidth}{%
\normalsize\sagesf\thepage\hfill\itshape\\[-6pt]
\noindent\rule{\textwidth}{0.25pt}}}
\def\@oddhead{\parbox{\textwidth}{%
\normalsize\sagesf{\itshape{\leftmark}}\hfill\thepage\\[-6pt]
\noindent\rule{\textwidth}{0.25pt}}}
\def\@evenfoot{}
\def\@oddfoot{}
}

\def\ps@title{%
\def\@oddhead{\parbox{\textwidth}{\mbox{}\\[-1pt]%
\noindent\rule{\textwidth}{0.5pt}%
}}%
\let\@evenhead\@oddhead
\def\@oddfoot{}
\let\@evenfoot\@oddfoot}
\makeatother

\begin{document}

\runninghead{Ticozzi et al.}

\title{State estimation for a hybrid rigid-continuum space manipulator system}

\author{Lorenzo Ticozzi\affilnum{1}, Patricio A.~Vela\affilnum{2}, and Panagiotis Tsiotras\affilnum{1}}

\affiliation{\affilnum{1}School of Aerospace Engineering, Georgia Institute of Technology\\
\affilnum{2}School of Electrical and Computer Engineering, Georgia Institute of Technology}

\corrauth{Lorenzo Ticozzi, School of Aerospace Engineering, Georgia Institute of Technology, Atlanta, GA, USA 30332.}

\email{lorenzo@gatech.edu}

\begin{abstract}
Spacecraft-mounted robotic manipulators can support long-term operability of orbiting assets by conducting tasks such as on-orbit assembly, repair, upgrade, and refueling.
In this context, the integration of continuum/soft robotic mechanisms into space manipulator designs promises to reduce stowage volume and mass, as well as to improve
operational safety through the inherent mechanical compliance of soft structures.
This work presents a novel state reconstruction pipeline for a hybrid soft-rigid space manipulator, in which the distal link is a tendon-driven continuum robot.
The estimation problem is formulated as a maximum a posteriori (MAP) inference problem over the hybrid manipulator's configuration, conditioned on discrete, noisy measurements from joint encoders and a vision-based sensing pipeline. 
The resulting soft-constrained optimization problem is transcribed as a factor graph, providing a compact representation for fusing heterogeneous measurements with the rigid-soft kinematic model.
Furthermore, the adoption of the Geometric Variable Strain (GVS) approach yields a minimum-order parameterization of the configuration, a property expected to facilitate model-based feedback control design in future work.
The proposed implementation is validated in simulation on a closed-loop trajectory where, for better realism, position measurements for the continuum link are obtained by means of a simulated visual pipeline leveraging synthetic photorealistic images.
The state reconstruction accuracy and computational performance confirm the potential of the proposed framework for hybrid soft-rigid robotic applications in orbit. 

\end{abstract}

\keywords{Space robotics, hybrid soft-rigid manipulators, factor graphs, continuum robot state estimation}

\maketitle

\section{Introduction}

\subsection{Related work}
Technological advances, together with the standardization of satellite subsystems, have democratized access to space and enabled large-scale satellite production and operation.
This trend drives sustained growth of the space economy, projected at an average annual rate of 9\%.~\cite{wef2024}
At the same time, the rapid expansion of the Low Earth Orbit (LEO) satellite population and the associated growth of space debris are challenging the current space exploitation model, motivating the search for a more sustainable paradigm.

Significant effort has therefore been devoted to developing robotic on-orbit servicing (OOS) capabilities, aimed at enabling value-added tasks such as on-orbit assembly, refueling, and maintenance.
Autonomous spacecraft-mounted manipulators are a key enabling technology for this vision, owing to their versatility across a wide range of OOS missions.~\cite{tsiotras2023}
Designing and operating such systems, however, remains challenging, as it must reconcile stringent accuracy and safety requirements with uncertain and dynamic operating conditions.

Historically, the majority of the literature on the subject has focused on modeling~\cite{dubowsky1993, valverde2018} and control design~\cite{senda1995, giordano2016, antonello2019} for rigid or rigid-flexible space manipulators.
More recently, however, soft and continuum robotics elements have been introduced into the field, motivated by the reduced mass and inertia afforded by tendon-driven architectures and by the inherent compliance of continuum manipulators.
Specifically, cable actuation for space manipulators was first investigated through the Tendon-Actuated Lightweight In-Space MANipulator (TALISMAN) prototype,~\cite{doggett2014, dorsey2015} followed by several studies addressing modeling and control design for tendon-actuated rigid arms.~\cite{li2019, kernot2022}
In parallel, the dynamical model was derived for the so-called Continuum-Manipulator Space Robot (CMSR),~\cite{yang2022} consisting of two tendon-driven continuum manipulators, modeled as Euler-Bernoulli beams, mounted on a satellite base; follow-up studies addressed control design for the CMSR.~\cite{yang2022jgcd, yang2024}
While these works pioneered spacecraft-mounted continuum manipulators, the CMSR dynamical model was formulated as a non-minimal set of differential-algebraic equations (DAEs), thereby complicating simulation, analysis, and control design.
Building upon the Geometric Variable Strain (GVS) formulation,~\cite{mathew2024} a later work addressed these limitations and introduced a minimal-order dynamical model for a hybrid soft-rigid space manipulator system.~\cite{ticozzi2025}

While the current state-of-the-art reflects a growing interest in leveraging soft robotics principles for OOS missions, several aspects of the problem remain underexplored.
In particular, existing state estimation methods for continuum robots are largely developed and validated under the assumption of a fixed base, with sensing relying on an external motion capture system or using fiber-embedded strain sensors.~\cite{lilge2022}
These assumptions are incompatible with platforms operating in space, where the base is a free-floating satellite bus whose own pose is uncertain, and where onboard sensing may be constrained to only vision and encoder measurements.
Conversely, conventional state estimation methodologies for rigid-body space manipulators have yet to be adapted to the case of continuum robotic links.
In particular, a probabilistic state reconstruction framework for the combined satellite-bus and continuum manipulator platform, jointly accounting for base and appendage uncertainty within a consistent kinematic and sensing model, has yet to be investigated.

\subsection{Contribution}
Motivated by the previous knowledge gap, this article presents a framework for the complete state estimation of a hybrid soft-rigid space manipulator system (HSMS).
The proposed contribution bridges the geometric HSMS model and recent factor graph-based problem formulations for continuum-robot state estimation,~\cite{ticozzi2025,ticozzi2026} with the following advantages.
First, the proposed approach is general and can be readily specialized to any soft-rigid kinematic chain and sensor configuration.
Second, it achieves sensor fusion at the system level: information is propagated across the entire kinematic chain, improving estimation accuracy and robustness under partial sensing.
Third, it is scalable and computationally efficient, a critical requirement for space-grade hardware.
The proposed framework is evaluated in simulation, with continuum-link position measurements generated via stereo triangulation using photorealistic synthetic images.

\section{Mathematical preliminaries}
\label{sec:theory}

\subsection{Hybrid soft-rigid kinematics}
\label{sec:kinematics}
\begin{figure}
\centering
\includegraphics[width=3.3in]{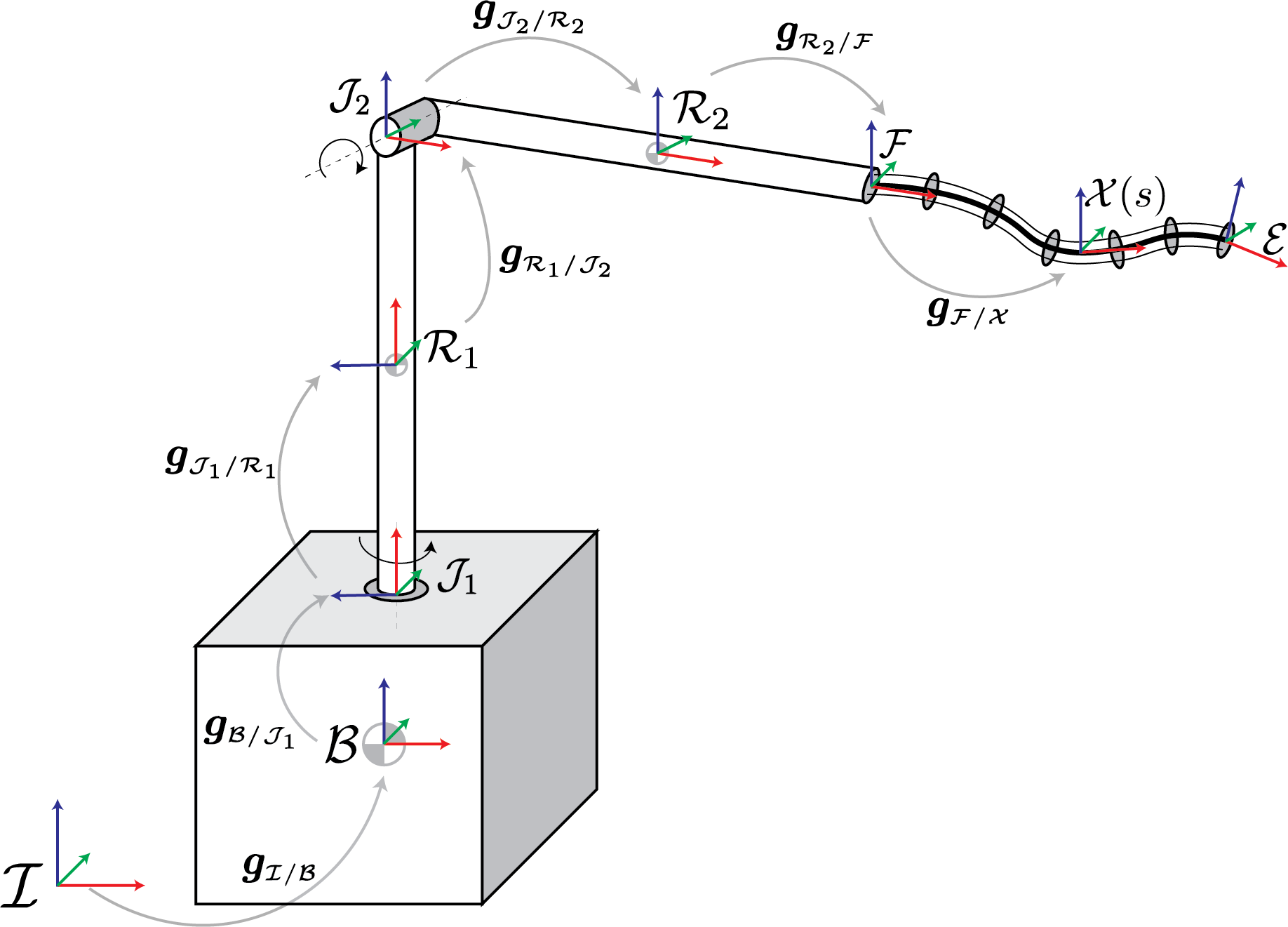}
\caption{Diagram of a hybrid space manipulator system. The hybrid soft-rigid manipulator is attached to the satellite bus and features two rigid links, actuated by revolute joints, and a tendon-driven continuum link.
Frame $\mathcal{I}$ can be interpreted, for instance, as the Hill frame centered on a client satellite.}
\label{fig:hsms-scheme}
\end{figure}

Throughout this study, the HSMS introduced in our previous work is considered~\cite{ticozzi2025}, as shown in Fig.~\ref{fig:hsms-scheme}, with forward kinematics from frame $\mathcal{I}$ to frame $\mathcal{F}$ as follows,
\begin{multline}
    \bm{g}_{\sss\mathcal{I} / \mathcal{F}}(\bm{q}_{\sss R}) 
    = 
    \bm{g}_{\sss\mathcal{I} / \mathcal{B}}(\bm{q}_{\sss \mathcal{B}})
    \bm{g}_{\sss\mathcal{B} / \mathcal{J}_1}(q_1)
    \bm{g}_{\sss\mathcal{J}_1 / \mathcal{R}_1}
    \bm{g}_{\sss\mathcal{R}_1 / \mathcal{J}_2}(q_2) \\
    \bm{g}_{\sss\mathcal{J}_2 / \mathcal{R}_2} 
    \bm{g}_{\sss\mathcal{R}_2 / \mathcal{F}}.
    \label{eq:fw_kin_rigid}
\end{multline}
Here, $\bm{g}_{\sss\mathcal{X} / \mathcal{Y}} \in \mathrm{SE}(3)$ denotes the pose of frame $\mathcal{Y}$ with respect to frame $\mathcal{X}$, $\bm{q}_{\sss \mathcal{B}} \in \mathbb{R}^6$ collects the orientation and translation coordinates of the base, $q_1$ and $q_2$ denote the angular displacements of the revolute joints 1 and 2, respectively.
The configuration of the rigid part of the system is denoted compactly by $\bm{q}_{\sss R} \in \mathbb{R}^8$,
\begin{equation}
    \bm{q}_{\sss R} \triangleq [\bm{q}_{\sss \mathcal{B}}^\top, q_1, q_2]^\top.
\end{equation}

The configuration of the continuum appendage is described by the curve $\bm{g}_{\sss\mathcal{F} / \mathcal{X}}(\cdot) : [0, L] \to \mathrm{SE}(3)$, parameterized by the arc-length coordinate $s$, 
\begin{equation}
    \label{eq:fw_kin_complete}
    \bm{g}_{\sss\mathcal{I} / \mathcal{E}} 
    = \bm{g}_{\sss\mathcal{I} / \mathcal{F}}(\bm{q}_{\sss R}) \bm{g}_{\sss\mathcal{F} / \mathcal{X}}(s=L) 
    = \bm{g}_{\sss\mathcal{I} / \mathcal{F}}(\bm{q}_{\sss R}) \bm{g}_{\sss\mathcal{F} / \mathcal{E}},
\end{equation}
where $\mathcal{E}$ is the end-effector frame.
Along the continuum appendage, the following linear differential relationship holds between the pose field $\bm{g}(s) \in \mathrm{SE}(3)$ and the strain field, $\bm{\xi}(s) \in \mathfrak{se}(3)$,
\begin{equation}
\label{eq:diff-kin}
    \bm{g}^\prime(s) = \bm{g}(s)\bm{\xi}(s),
\end{equation}
where $\bm{g}(0) = \bm{I}_4$ and $(\cdot)^\prime \triangleq \partial / \partial s (\cdot)$.~\cite{renda2020}
Notably, Eq.~\eqref{eq:diff-kin} admits a solution of the form,~\cite{magnus1954}
\begin{equation}
    \label{eq:g_prime_sol}
    \bm{g}(s) = \exp{\big(\bm{\Omega}(s)\big)},
\end{equation}
where $\bm{\Omega}(s) \in \mathfrak{se}(3)$ is given by the Magnus expansion of the strain field.~\cite{blanes1998}

The problem posed by Eq.~\eqref{eq:diff-kin} can be reduced to a finite-dimensional one through the Geometric Variable Strain (GVS) formulation,~\cite{mathew2024} which projects the strain field $\bm{\xi}(s)$ onto a finite set of basis functions $\bm{\Phi}_{\bm{\xi}}$, as follows,
\begin{equation}
    \label{eq:gvs}
    \bm{\xi}(s ; \bm{q}_{\sss S}) = \bm{\Phi}_{\bm{\xi}}(s)\bm{q}_{\sss S} + \bm{\xi}^*,
\end{equation}
where $\bm{q}_{\sss S} \in \mathbb{R}^S$ is a vector of generalized strain coordinates and $\bm{\xi}^* \in \mathfrak{se}(3)$ denotes the reference strain state.
Therefore, the configuration of the HSMS is fully described by the vector,
\begin{equation}
    \label{eq:q}
    \bm{q} \triangleq \left[\bm{q}_{\sss R}^\top, \bm{q}_{\sss S}^\top\right]^\top \in \mathbb{R}^{8+S}.
\end{equation}

\subsection{Factor graph inference}

\begin{figure}
\centering
\includegraphics[width=2.5in]{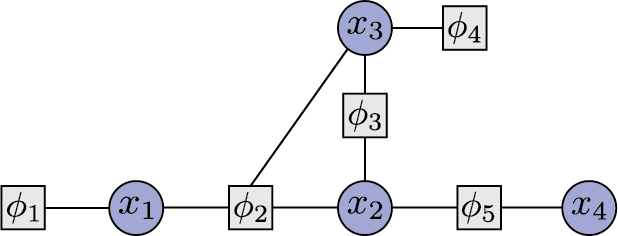}
\caption{Factor graph example. Variable nodes are represented as circles, while factor nodes are represented as squares.}
\label{fig:fg-example}
\end{figure}

Factor graphs are bipartite graphs whose nodes are of two types, factor nodes and variable nodes, with each edge connecting nodes of different types.
Factor graphs can represent any function $\phi$ that factors over a set of variables $x$ as
\begin{equation}
    \phi(x) = \prod_i \phi_i(x_i),
\end{equation}
where $x_i \subseteq x$ is the subset of variables adjacent to factor $\phi_i$ in the graph.
For example, the factor graph in Fig.~\ref{fig:fg-example} encodes the factored function
$\phi(x) = \phi_1(x_1)
\phi_2(x_1,x_2,x_3)
\phi_3(x_2,x_3)
\phi_4(x_3)
\phi_5(x_4)$.
This structure makes factor graphs a natural representation for the factored form of joint posterior distributions.~\cite{dellaert2017}
Given an unknown state $q$ and the corresponding observations $z$, by Bayes' rule, $p(q | z) \propto p(q)\, p(z | q)$, and the maximum a posteriori (MAP) estimate is
\begin{equation}
\label{eq:map1}
     q^{\mathrm{MAP}} = \argmax_{q}\, p(q)\, p(z | q),
\end{equation}
whose right-hand side can be represented by the corresponding factor graph.

\section{Problem formulation}
In principle, reconstructing the configuration of the HSMS in Fig.~\ref{fig:hsms-scheme} amounts to estimating the vector $\bm{q}$ defined in Eq.~\eqref{eq:q}.
However, the absolute poses $\bm{g}_{\sss\mathcal{I} / \mathcal{B}}, \bm{g}_{\sss\mathcal{I} / \mathcal{J}_1}, \ldots, \bm{g}_{\sss\mathcal{I} / \mathcal{R}_2}$ are also treated as estimation variables, for reasons detailed in the next section.
Moreover, $N+1$ pose variables are defined along the continuum appendage, $\bm{g}_i \triangleq \bm{g}_{\sss\mathcal{I}/\mathcal{X}}(s_i)$, for $i=0,\ldots,N$, with $s_0=0$ and $s_N=L$.~\cite{ticozzi2026} 
By construction, $\bm{g}_0 \equiv \bm{g}_{\sss\mathcal{I} / \mathcal{F}}$ and $\bm{g}_N \equiv \bm{g}_{\sss\mathcal{I} / \mathcal{E}}$.
Collecting these, and with a slight abuse of notation, the vector of unknowns is defined as
\begin{equation}
    \bm{\theta} 
    \triangleq 
    \{
    \bm{g}_{\sss\mathcal{I} / \mathcal{B}},
    \bm{g}_{\sss\mathcal{I} / \mathcal{J}_1},
    \ldots,
    \bm{g}_{\sss\mathcal{I} / \mathcal{R}_2},
    \bm{g}_0,
    \bm{g}_1,
    \ldots,
    \bm{g}_N,
    \bm{q}
    \}.
\end{equation}
Finally, recalling Eq.~\eqref{eq:map1}, the MAP estimate of the HSMS configuration is
\begin{equation}
\label{eq:map-2}
    \bm{\theta}^{\mathrm{MAP}}
    =
    \argmax_{\bm{\theta}}
    p(\bm{\theta})
    p(\bm{z} | \bm{\theta}),
\end{equation}
where $\bm{z}$ collects all available sensor measurements (e.g., from star trackers, joint encoders, and strain sensors).

\subsection{Hybrid space manipulator factor graph}
\label{sec:hybrid-fg}
\begin{figure*}
\centering
\includegraphics[]{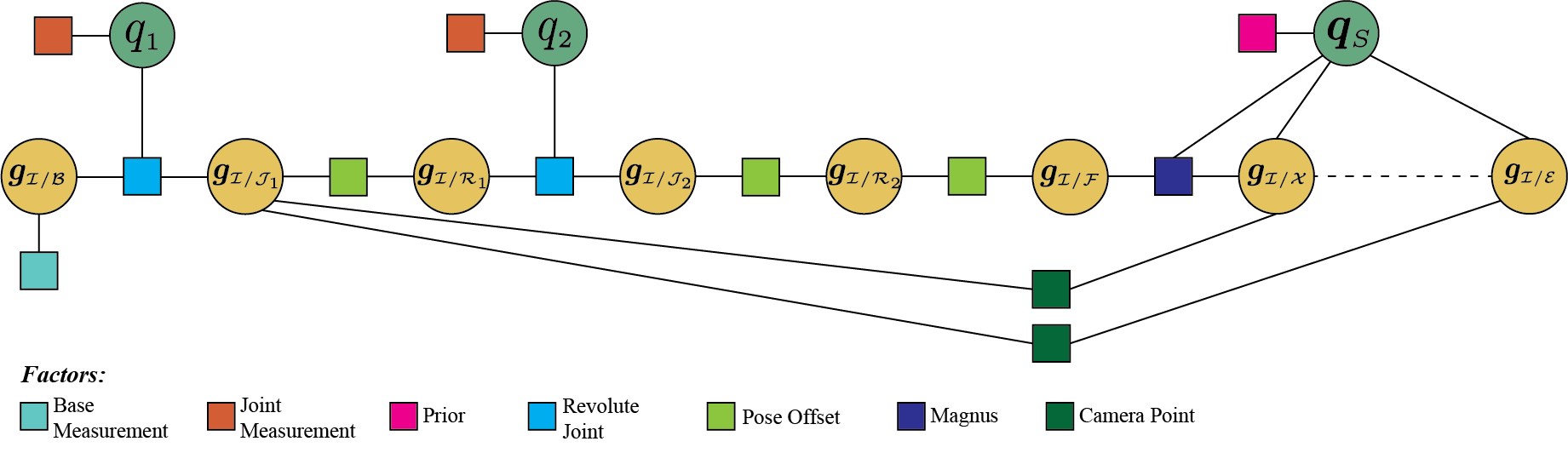}
\caption{Hybrid space manipulator factor graph.
Variables are represented by circles (yellow for poses, green for scalars and vectors); factor nodes are represented by squares.}
\label{fig:hybrid-fg}
\end{figure*}
Given the system kinematics and the estimation variables defined in the previous sections, the joint posterior $p(\bm{\theta} | \bm{z})$ factors into a set of relations that expose the local dependency structure of the variables of the problem.
This structure is made explicit by the factor graph in Fig.~\ref{fig:hybrid-fg}, whose factors are described in the following.

\subsubsection{Base measurement factor:}
The root of the kinematic chain (frame $\mathcal{B}$ in Fig.~\ref{fig:hsms-scheme}) is related to frame $\mathcal{I}$ by a noisy measurement of the base pose;
for simplicity, we assume that a relative navigation stack is available that measures the relative pose of $\mathcal{B}$ with respect to $\mathcal{I}$.
The output of the navigation module is denoted by $\tilde{\bm{g}}_{\sss \mathcal{I} / \mathcal{B}}$,
\begin{equation}
\label{eq:pose-meas}
    \tilde{\bm{g}}_{\sss \mathcal{I} / \mathcal{B}} = \bm{g}_{\sss \mathcal{I} / \mathcal{B}} \exp{(\bm{w}_{ \mathcal{B}})},
\end{equation}
where $\bm{w}_{\mathcal{B}} \in \mathfrak{se}(3)$ is zero-mean Gaussian noise with standard deviations $\sigma_{\mathcal{B}, \mathrm{rot}}$ and $\sigma_{\mathcal{B}, \mathrm{pos}}$ for the orientation and position, respectively.
The associated unary residual is defined as
\begin{equation}
\label{eq:base-meas}
    \bm{e}_{\mathcal{B}}
    \bigl(
    \bm{g}_{\sss\mathcal{I} / \mathcal{B}}
    \bigr) 
    =
    \log{\bigl(
    \tilde{\bm{g}}_{\sss \mathcal{I} / \mathcal{B}}^{-1}
    \bm{g}_{\sss\mathcal{I} / \mathcal{B}}
    \bigr)
    },
\end{equation}
which vanishes in the ideal case where $\tilde{\bm{g}}_{\sss \mathcal{I} / \mathcal{B}} = \bm{g}_{\sss \mathcal{I} / \mathcal{B}}$.
The base measurement factor (represented by the turquoise square in Fig.~\ref{fig:hybrid-fg}) is then modeled as a Gaussian potential,
\begin{equation}
\label{eq:base-factor}
    \phi_{ \mathcal{B}}
    \bigl(
    \bm{g}_{\sss\mathcal{I} / \mathcal{B}}
    \bigr)
    \propto
    \exp{\biggl(-\frac{1}{2} \left\| \bm{e}_{\mathcal{B}}
    \right\|_{\bm{\Sigma}_{\mathcal{B}}}^2\biggr)},
\end{equation}
where $\left\| \bm{e} \right\|_{\bm{\Sigma}}^2=\bm{e}^\top \bm{\Sigma}^{-1} \bm{e}$ is the squared Mahalanobis distance.
To maintain a safety margin, the covariance $\bm{\Sigma}_{\mathcal{B}}$ is defined by inflating the sensor standard deviations by a factor of $\sqrt{3}$, as follows,
\begin{equation}
\label{eq:Sigma_B}
    \bm{\Sigma}_{\mathcal{B}}
    = 
    \begin{bmatrix}
        3 \sigma_{\mathcal{B}, \mathrm{rot}}^2 \bm{I}_3 & \bm{0}  \\
        \bm{0}   & 3 \sigma_{\mathcal{B}, \mathrm{pos}}^2 \bm{I}_3
    \end{bmatrix}.
\end{equation}
\subsubsection{Pose offset factors:}
Consider the known offset $\bm{g}_{\sss \mathcal{R}_2 / \mathcal{F}}$ between frames $\mathcal{R}_2$ and $\mathcal{F}$.
Accordingly, the kinematic relationship is described by the residual
\begin{equation}
    \bm{e}^{\mathrm{off}}_{\sss \mathcal{R}_2 / \mathcal{F}}
    \bigl(
    \bm{g}_{\sss \mathcal{I} / \mathcal{R}_2},
    \bm{g}_{\sss \mathcal{I}/\mathcal{F}}
    \bigr)
    =
    \log{
    \bigl(
    \bm{g}_{\sss \mathcal{R}_2 / \mathcal{F}}^{-1}
    \bm{g}_{\sss \mathcal{I} / \mathcal{R}_2}^{-1}
    \bm{g}_{\sss \mathcal{I}/\mathcal{F}}
    \bigr)
    },
\end{equation}
and by the corresponding factor (light green squares in Fig.~\ref{fig:hybrid-fg}),
\begin{equation}
\phi^{\mathrm{off}}_{\sss \mathcal{R}_2 / \mathcal{F}}
    \bigl(
    \bm{g}_{\sss \mathcal{I} / \mathcal{R}_2},
    \bm{g}_{\sss \mathcal{I}/\mathcal{F}}
    \bigr)
    \propto
    \exp{\biggl(-\frac{1}{2} \left\| \bm{e}^{\mathrm{off}}_{\sss \mathcal{R}_2 / \mathcal{F}}
    \right\|_{\bm{\Sigma}_{\mathrm{off}}}^2\biggr)}.
\end{equation}
While the exact pose composition could be used to obtain $\bm{g}_{\sss\mathcal{I}/\mathcal{F}}$ directly from $\bm{g}_{\sss\mathcal{I}/\mathcal{R}_2}$ and $\bm{g}_{\sss\mathcal{R}_2/\mathcal{F}}$, encoding such offsets through factors has two main advantages.
First, the resulting formulation is agnostic to the sensing configuration.
For instance, additional visual information on a rigid link is processed by a factor on the corresponding pose variable, leaving the remainder of the graph unaltered.
Second, optimizing the factor graph also yields a measure of confidence associated with each estimated variable, crucial in safety-critical OOS scenarios.

\subsubsection{Revolute joint factors:}
Revolute joint factors (cyan squares in Fig.~\ref{fig:hybrid-fg}) encode the transformation between the absolute pose of a parent body frame and that of the child joint frame.
For the first joint, this reads
\begin{equation}
\label{eq:joint-kin}
    \bm{g}_{\sss\mathcal{I} / \mathcal{J}_1}
    =
    \bm{g}_{\sss\mathcal{I} / \mathcal{B}}
    \bm{g}_{\sss \mathcal{B} / \mathcal{J}_1}^{\mathrm{off}}
    \exp{(\bm{\xi}_1 q_1)},
\end{equation}
where $\bm{g}_{\sss \mathcal{B} / \mathcal{J}_1}^{\mathrm{off}} \in \mathrm{SE}(3)$ is a fixed, known offset from the parent frame to the joint frame in its home configuration ($q_1=0$), and $\bm{\xi}_1 \in \mathfrak{se}(3)$ is the unit joint twist expressed in the joint frame.
From Eq.~\eqref{eq:joint-kin}, the residual vector is defined as
\begin{multline}
    \bm{e}_{\sss \mathcal{J}_1}^R
    \bigl(
    \bm{g}_{\sss\mathcal{I} / \mathcal{B}},
    \bm{g}_{\sss\mathcal{I} / \mathcal{J}_1},
    q_1
    \bigr)
    = \\
    \log \Bigl(
    \exp \bigl(
    -\bm{\xi}_1 q_1
    \bigr)
    (\bm{g}_{\sss \mathcal{B} / \mathcal{J}_1}^{\mathrm{off}})^{-1}\,
    \bm{g}_{\sss\mathcal{I} / \mathcal{B}}^{-1}
    \bm{g}_{\sss\mathcal{I} / \mathcal{J}_1}
    \Bigr),
\end{multline}
with corresponding factor,
\begin{equation}
    \phi_{\sss \mathcal{J}_1}^R
    \bigl(
    \bm{g}_{\sss\mathcal{I} / \mathcal{B}},
    \bm{g}_{\sss\mathcal{I} / \mathcal{J}_1},
    q_1
    \bigr)
    \propto
    \exp{
    \biggl(-\frac{1}{2} \left\| 
    \bm{e}_{\sss \mathcal{J}_1}^R
\right\|_{\bm{\Sigma}_{\mathcal{J}}^R}^2\biggr)}.
\end{equation}
The same construction yields $\phi_{\sss \mathcal{J}_2}^R (\bm{g}_{\sss\mathcal{I} / \mathcal{R}_1}, \bm{g}_{\sss\mathcal{I} / \mathcal{J}_2}, q_2)$ for the second revolute joint, and extends to other joint types (e.g., prismatic), providing a general family of kinematic factors that encode the pose relationships along a rigid multibody chain.

\subsubsection{Joint measurement factors:}
Joint encoder measurements are modeled as
\begin{equation}
\label{eq:encod-meas}
    \tilde{q}_j = q_j + w_{q},
\end{equation}
where $\tilde{q}_j \in \mathbb{R}$ denotes the $j$-th encoder reading and $w_q$ is zero-mean Gaussian noise with standard deviation $\sigma_q$. 
The associated scalar residual is given by
\begin{equation}
    e_j^{q}(q_j)
    =
    q_j - \tilde{q}_j,
\end{equation}
yielding the measurement factor
\begin{equation}
    \phi_j^q(q_j)
    =
    \exp{\biggl(-\frac{1}{2} \left\| 
    e_{j}^q
\right\|_{\Sigma_{q}}^2\biggr)},
\end{equation}
represented by the orange squares in Fig.~\ref{fig:hybrid-fg} for $j=1,2$.
Similarly to Eq.~\eqref{eq:Sigma_B}, the covariance $\Sigma_q$ is assigned as $\Sigma_q = 3\sigma_q^2$.

\begin{figure}
\centering
\includegraphics[width=2.5in]{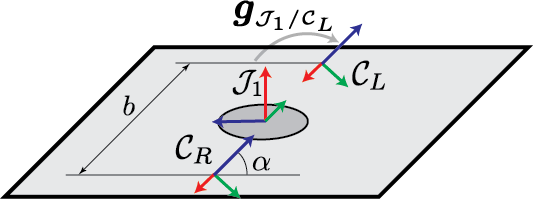}
\caption{Upper face of the HSMS base with the left and right camera frames, denoted by $\mathcal{C}_L$ and $\mathcal{C}_R$, respectively.
The camera elevation angle and stereo baseline are denoted by $\alpha$ and $b$, respectively, with $\alpha=41.6^\circ$ and $b=0.4$ m.
}
\label{fig:camera-scheme}
\end{figure}
\subsubsection{Camera point factors:}
A vision-based approach is adopted to reconstruct the configuration of the continuum link.
Two cameras in a stereo configuration are mounted on a rig which is rigidly attached to the joint frame $\mathcal{J}_1$, with parallel boresight axes directed towards the workspace of the continuum manipulator, according to the scheme in Fig.~\ref{fig:camera-scheme}.
Stereo triangulation yields three-dimensional measurements $\tilde{\bm{p}}_i^{\mathcal{C}_L}$ expressed in the left camera frame; at arclength $s_i$, the following measurement model holds,
\begin{equation}
    \bm{p}_i^{ \mathcal{C}_L} 
    = 
    t
    \bigl(
    \bm{g}_{\sss \mathcal{J}_1 / \mathcal{C}_L}^{-1}
    \bm{g}_{\sss \mathcal{I} / \mathcal{J}_1}^{-1}
    \bm{g}_i
    \bigr),
\end{equation}
where $t(\cdot) : \mathrm{SE}(3) \to \mathbb{R}^3$ extracts the translational part of a pose and $\bm{g}_{\sss \mathcal{J}_1 / \mathcal{C}_L}$ is the fixed offset between the joint frame and the camera frame.
The associated residual reads
\begin{equation}
\label{eq:cam-residual}
    \bm{e}_i^{ \mathcal{C}_L}
    \bigl(
    \bm{g}_{\sss \mathcal{I} / \mathcal{J}_1}, \bm{g}_i
    \bigr)
    =
    t
    \bigl(
    \bm{g}_{\sss \mathcal{J}_1 / \mathcal{C}_L}^{-1}
    \bm{g}_{\sss \mathcal{I} / \mathcal{J}_1}^{-1}
    \bm{g}_i
    \bigr) -
    \tilde{\bm{p}}_i^{ \mathcal{C}_L},
\end{equation}
where the argument of $t(\cdot)$ is the pose of the $i$-th backbone node relative to the camera.
The camera point factors (dark green squares in Fig.~\ref{fig:hybrid-fg}) are then defined as
\begin{equation}
    \phi_i^{ \mathcal{C}_L}
    \bigl(
    \bm{g}_{\sss \mathcal{I} / \mathcal{J}_1}, \bm{g}_i
    \bigr)
    \propto
    \exp{\biggl(-\frac{1}{2} 
    \left\| 
    \bm{e}_{i}^{\mathcal{C}_L}
\right\|_{\bm{\Sigma}_{\mathcal{C}_L}}^2
\biggr)}.
\end{equation}

\subsubsection{Continuum manipulator factors:}
Two factor types are specific to the continuum link: Magnus factors and the prior factor on the GVS coordinates $\bm{q}_{\sss S}$ (blue and magenta squares in Fig.~\ref{fig:hybrid-fg}, respectively).~\cite{ticozzi2026}
The former encodes the kinematic relation between consecutive backbone poses under the strain parameterization of Eq.~\eqref{eq:gvs} through the residual
\begin{equation}
    \label{eq:residual-Mag}
    \bm{e}_{i}^M
    \bigl(
    \bm{g}_i,\bm{g}_{i+1},\bm{q}_{\sss S}
    \bigr) 
    = 
    \log \Bigl(
    \exp \bigl(
    -\bm{\Omega}_i^{i+1}(\bm{q}_{\sss S})
    \bigr)
    \bm{g}^{-1}_i\,\bm{g}_{i+1}
    \Bigr),
\end{equation}
while the latter centers the strain coefficients at the origin, $\bm{e}^p(\bm{q}_{\sss S}) = \bm{q}_{\sss S}$.
Hereafter, these factors will be denoted by $\phi_i^M(\bm{g}_i,\bm{g}_{i+1},\bm{q}_{\sss S})$ and $\phi_p(\bm{q}_{\sss S})$, respectively.

\subsection{Factor graph optimization}
Given the factors defined above, the negative logarithm of their product transforms the MAP problem in Eq.~\eqref{eq:map-2} into an equivalent nonlinear least-squares (NLLS)
problem, as follows,
\begin{multline}
\label{eq:nlls}
    \bm{\theta}^{\mathrm{MAP}}
    =
    \argmin_{\bm{\theta}}
    \left\| 
    \bm{e}_{\mathcal{B}}
    \right\|_{\bm{\Sigma}_{\mathcal{B}}}^2
    +
    \left\|
    \bm{e}_p
    \right\|_{\bm{\Sigma}_{p}}^2
    + 
    \sum_{j=1}^2
    \Bigl(
    \bigl\|
    \bm{e}_{\mathcal{J}_j}^R
    \bigr\|_{\bm{\Sigma}_{\mathcal{J}}^R}^2
    \\
    +
    \bigl\|
    e_j^q
    \bigr\|^2_{\Sigma_q}
    \Bigr)
    +
    \sum_{(\mathcal{Y}, \mathcal{Z}) \in \mathcal{O}}
    \left\|
    \bm{e}_{\sss \mathcal{Y}/\mathcal{Z}}^{\mathrm{off}}
    \right\|_{\bm{\Sigma}_{\mathrm{off}}}^2
    +
    \sum_{i=0}^{N-1}
    \left\|
    \bm{e}_i^M
    \right\|_{\bm{\Sigma}_M}^2 
    \\
    +
    \sum_{i \in \mathcal{S}_C}
    \left\|
    \bm{e}_i^{\mathcal{C}_L}
    \right\|_{\bm{\Sigma}_{\mathcal{C}_L}}^2.
\end{multline}
In Eq.~\eqref{eq:nlls}, the set $\mathcal{O} = \{(\mathcal{J}_1,\mathcal{R}_1), (\mathcal{J}_2,\mathcal{R}_2),
(\mathcal{R}_2,\mathcal{F})\}$ collects the frame pairs connected by a constant pose offset, while $\mathcal{S}_C$ is the index set of the backbone nodes whose position is successfully triangulated.
The NLLS is solved iteratively through the Levenberg-Marquardt (LM) algorithm, with efficient iterations enabled by the analytic Jacobians of each residual with respect to the variables it is adjacent to.

The MAP solution of Eq.~\eqref{eq:nlls} corresponds to the HSMS configuration at a fixed time instant, since no notion of time variation is embedded in the proposed formulation.
Estimation of the HSMS configuration along a trajectory is therefore obtained by solving Eq.~\eqref{eq:nlls} repeatedly, seeding the LM solver at each time step with the solution retrieved at the previous one.

\section{Simulations}

\subsection{Data generation pipeline}
\begin{figure}[t]
\centering
\begin{subfigure}[b]{0.49\linewidth}
    \centering
    \includegraphics[width=\linewidth]{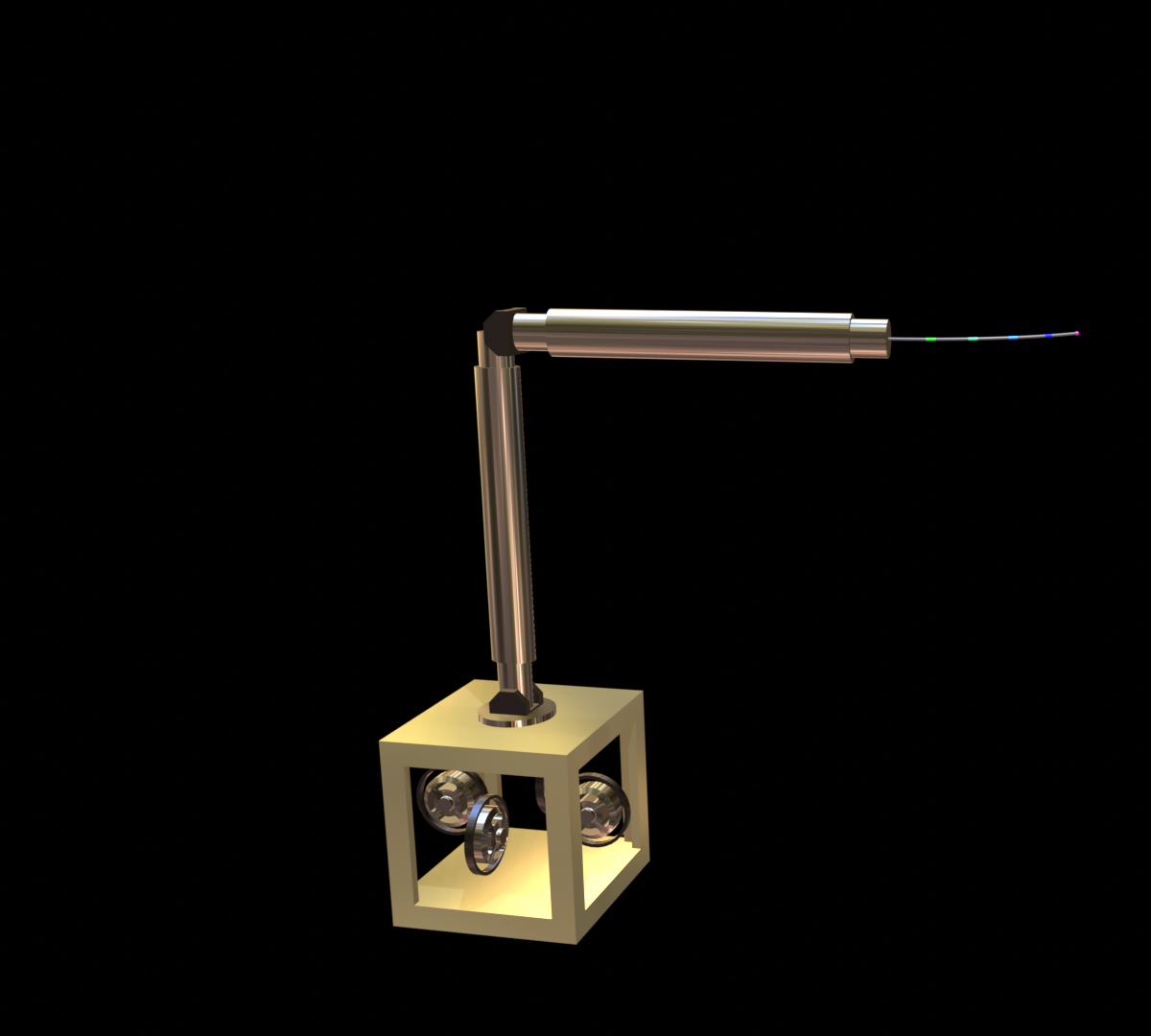}
    \caption{Configuration at $t=0$ s.}
    \label{fig:sub1}
\end{subfigure}
\hfill
\begin{subfigure}[b]{0.49\linewidth}
    \centering
    \includegraphics[width=\linewidth]{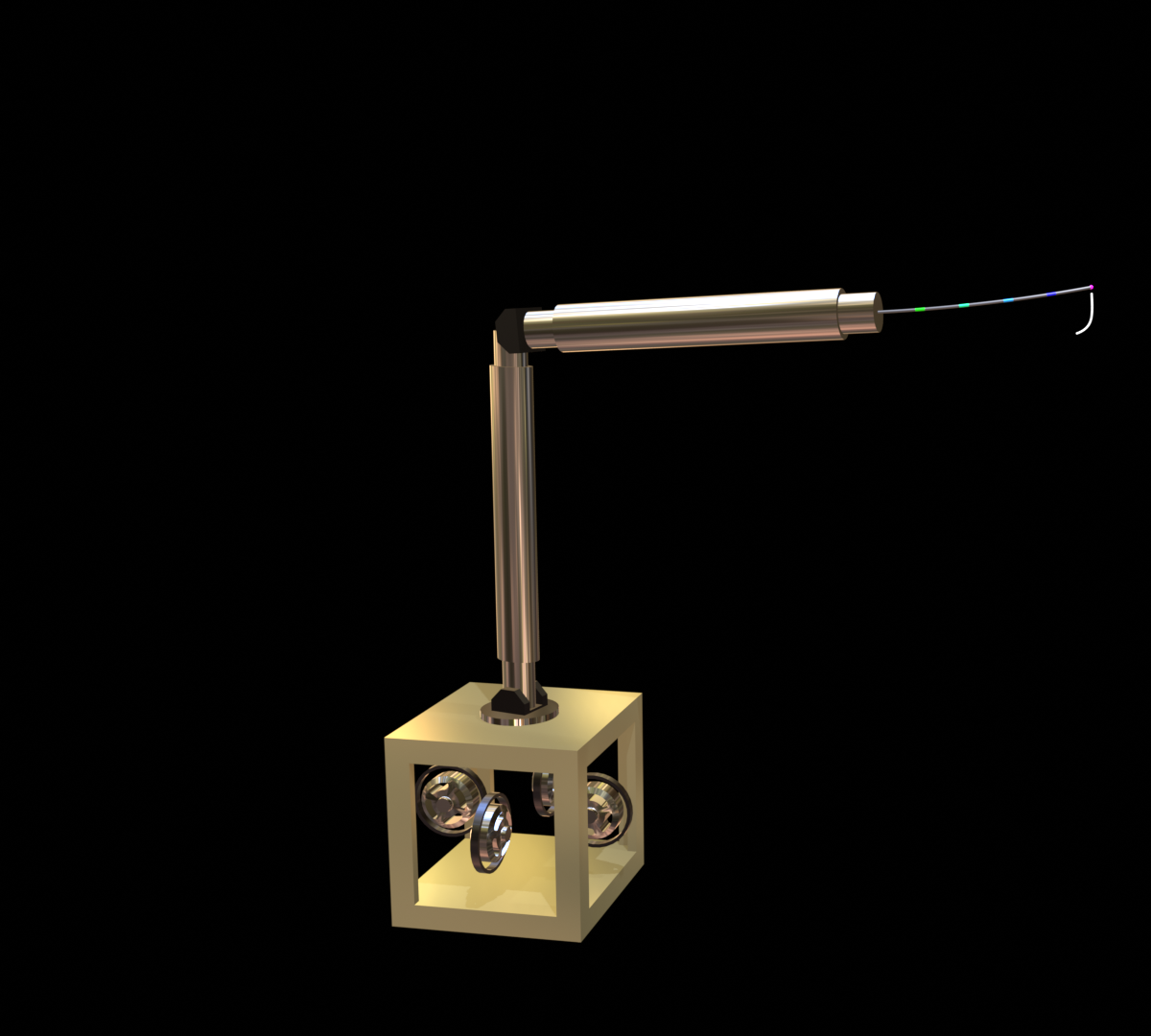}
    \caption{Configuration at $t=20$ s.}
    \label{fig:sub2}
\end{subfigure}

\vspace{0.5em}

\begin{subfigure}[b]{0.49\linewidth}
    \centering
    \includegraphics[width=\linewidth]{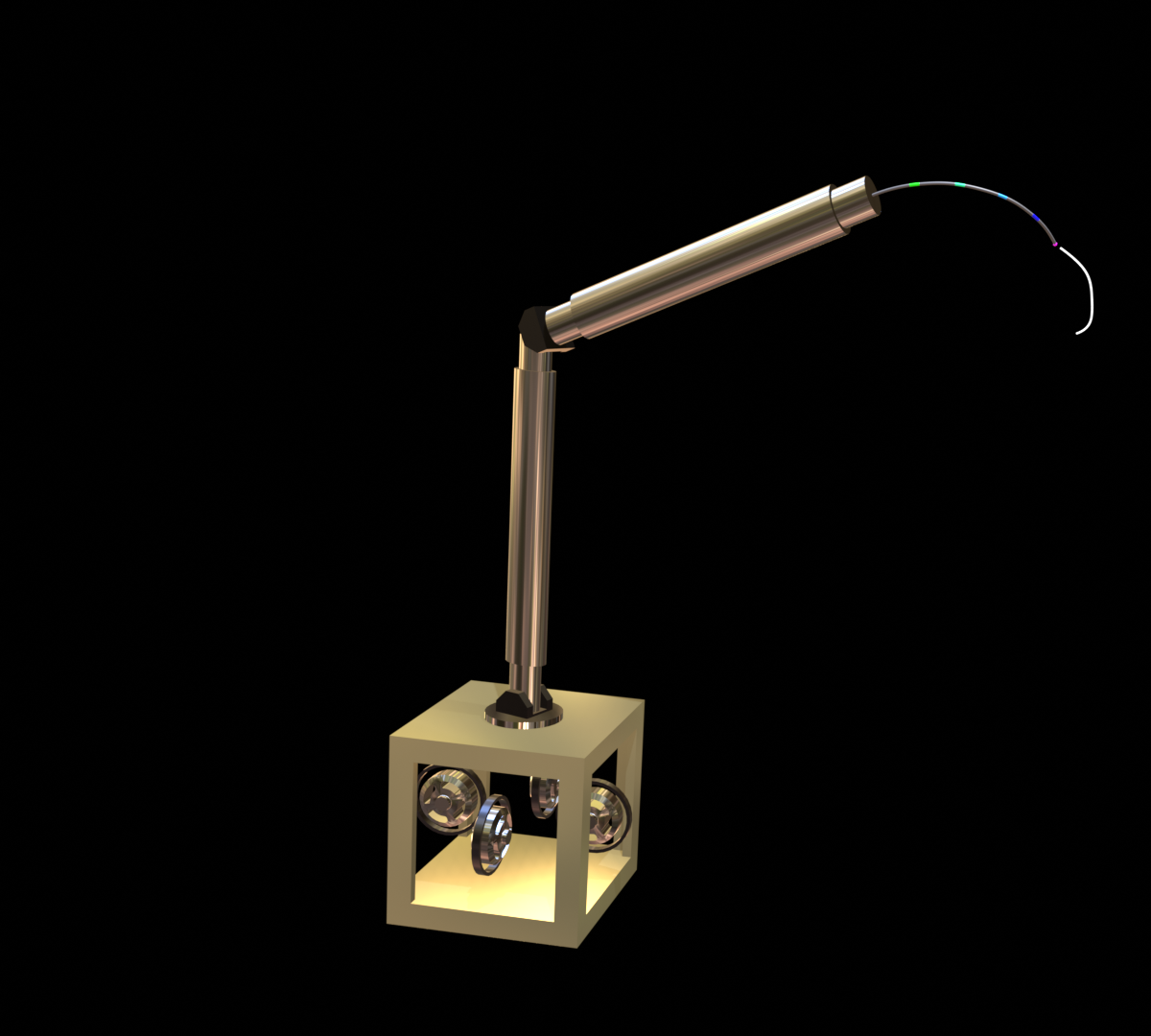}
    \caption{Configuration at $t=40$ s.}
    \label{fig:sub3}
\end{subfigure}
\hfill
\begin{subfigure}[b]{0.49\linewidth}
    \centering
    \includegraphics[width=\linewidth]{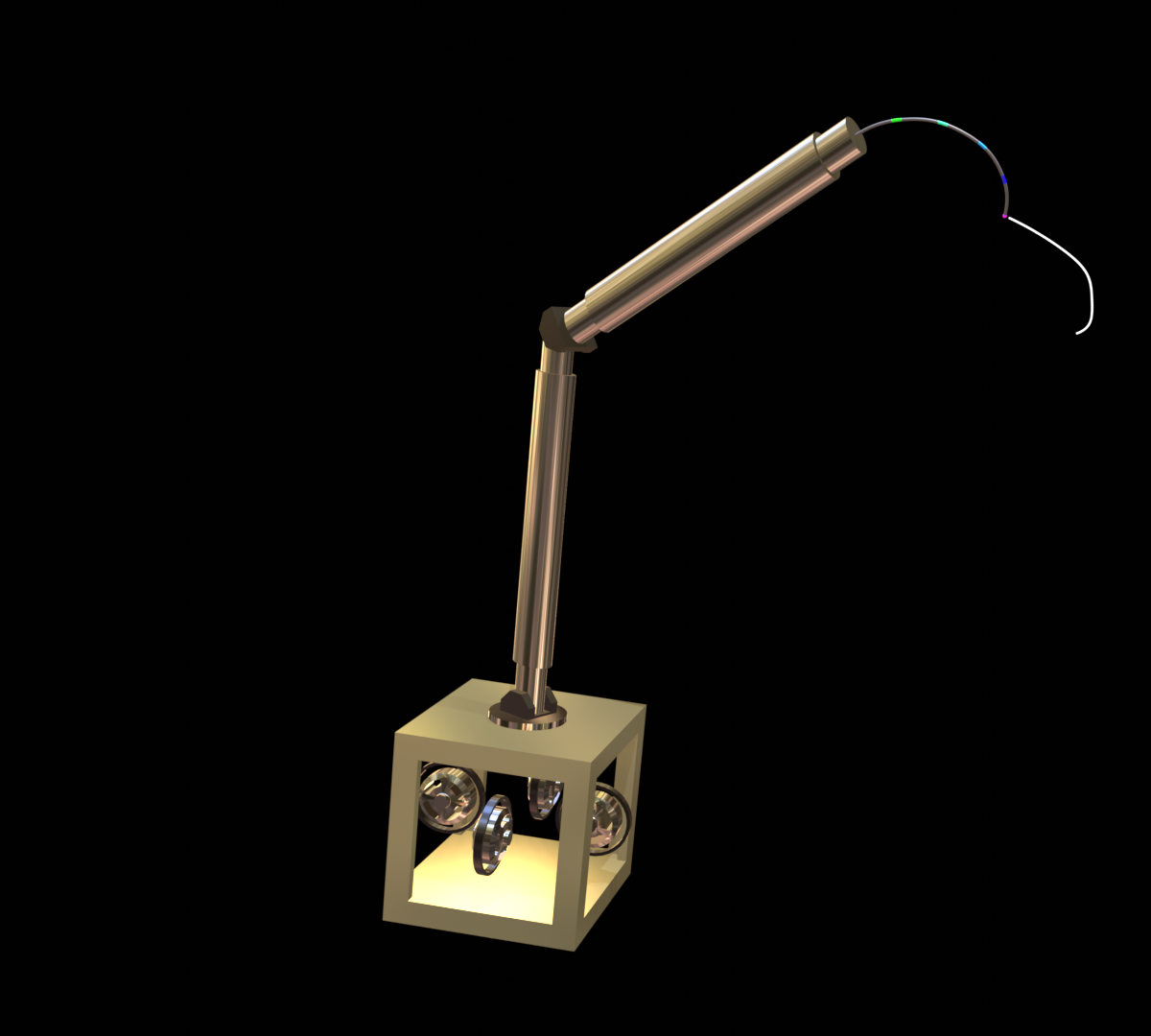}
    \caption{Configuration at $t=60$ s.}
    \label{fig:sub4}
\end{subfigure}
\caption{HSMS configuration along the $60$-s simulated maneuver, with the end-effector trajectory displayed in white.}
\label{fig:grid}
\end{figure}

Ground-truth data for the HSMS shown in Fig.~\ref{fig:hsms-scheme} was generated using the SoRoSim V6.3 library,~\cite{mathew2023} employing the HSMS model and control architecture detailed in our previous work.~\cite{ticozzi2025} 
The simulation executed a $60$-sec trajectory where the end-effector tracked a slowly evolving Cartesian position reference, with evolution displayed in Fig.~\ref{fig:grid}. 
Synthetic sensor data was generated by corrupting the ground truth with Gaussian noise.

Base pose measurements (Eq.~\eqref{eq:pose-meas}) used standard deviations $\sigma_{\mathcal{B},\mathrm{rot}} = 0.0175$ rad and $\sigma_{\mathcal{B},\mathrm{pos}} = 0.01$ m, reflecting standard relative navigation accuracies.~\cite{opromolla2017}
Similarly, joint encoder readings (Eq.~\eqref{eq:encod-meas}) applied a standard deviation of $\sigma_q = 10^{-4}$~rad.
\begin{figure}
\centering
\includegraphics[width=\linewidth]{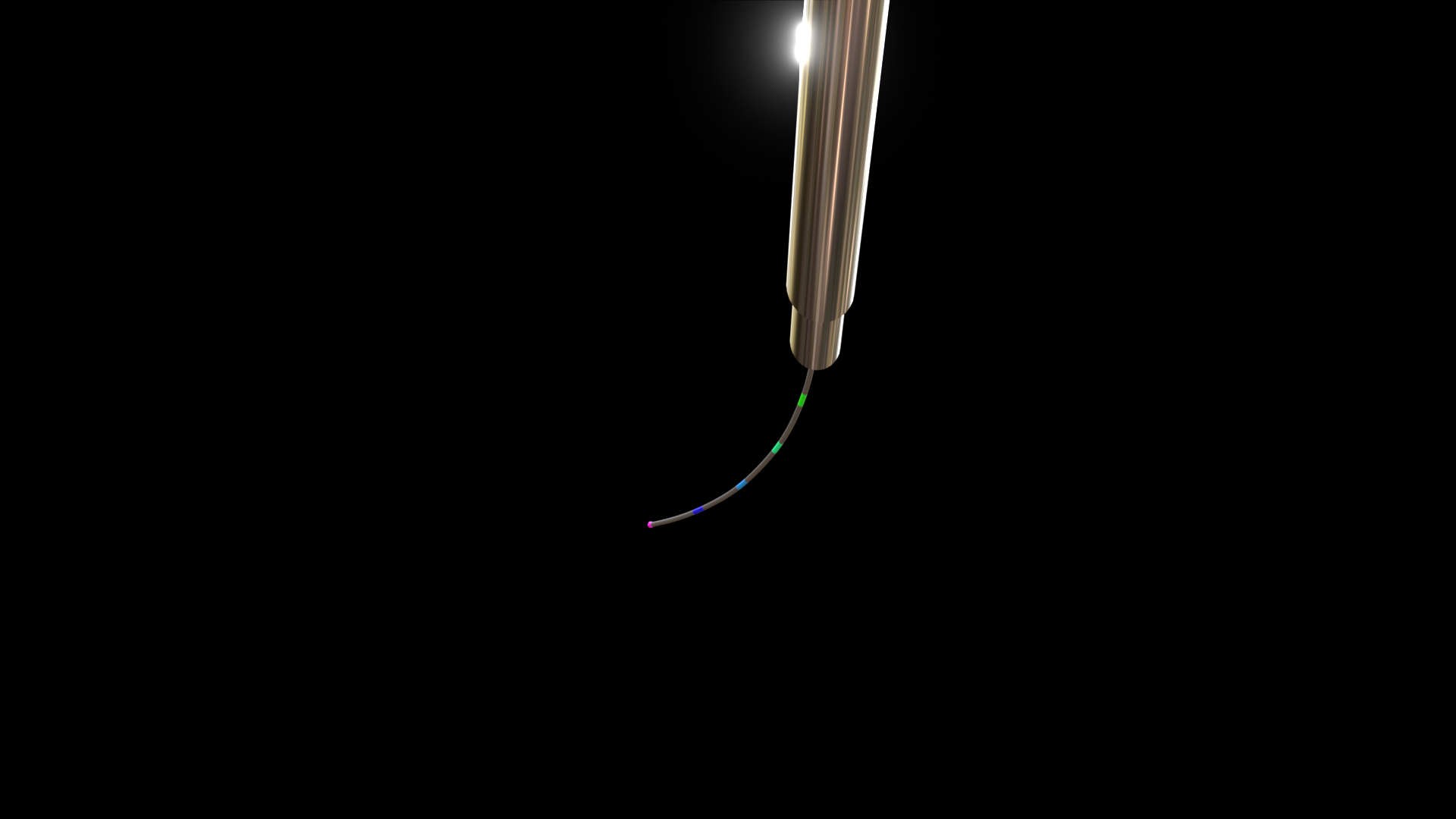}
\caption{Blender render of the left-camera view.
The continuum link carries five colored bands at equally-spaced arclength coordinates, which are
detected in both views and triangulated to obtain the position measurements
$\tilde{\bm{p}}_i^{\mathcal{C}_L}$ in Eq.~\eqref{eq:cam-residual}.}
\label{fig:left-view}
\end{figure}
\begin{figure*}[t!]
\centering
\includegraphics[width=\linewidth]{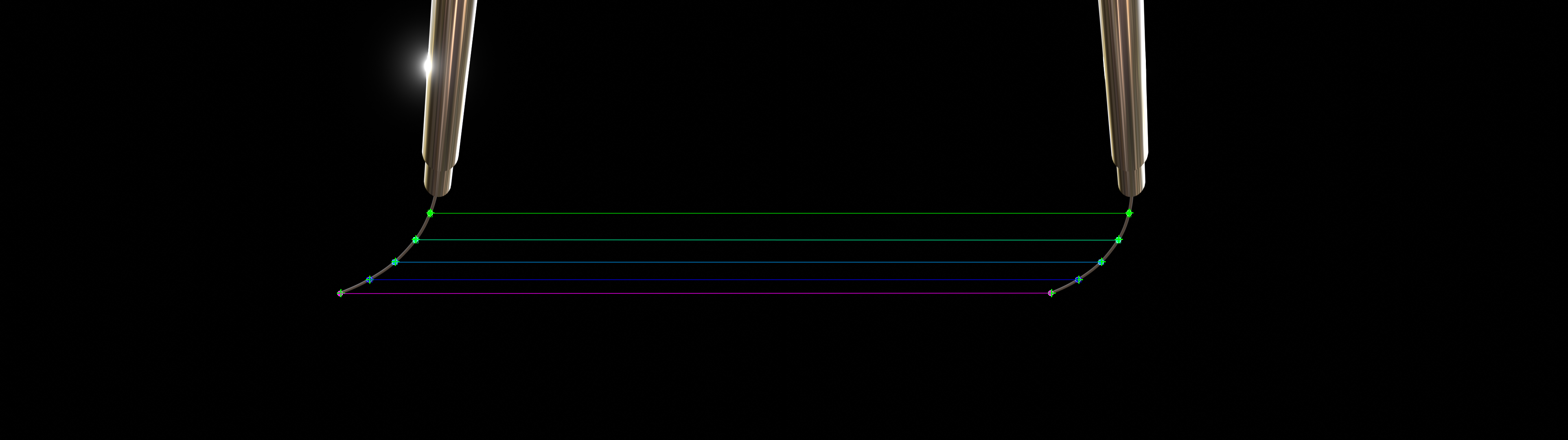}
\caption{Color-based stereo matching across the left and right images.}
\label{fig:stereo-match}
\end{figure*}
Synthetic images for the left and right cameras, whose configuration is shown in Fig.~\ref{fig:camera-scheme}, were rendered via a Blender model of the HSMS (see Fig.~\ref{fig:left-view}). 
Since the stereo rig is rigidly attached to $\mathcal{J}_1$, the relative geometry between the cameras and the manipulator is independent of $q_1$.

The cameras were modeled as ideal pinhole sensors with a $1920 \times 1080$ pixel resolution and a $61.9^\circ$ horizontal FOV, corresponding to a focal length of $1600$ pixels, so as to frame the continuum appendage at a nominal camera-to-appendage distance of approximately $3$~m.

\subsection{Visual frontend}

A marker-based visual frontend handles stereo triangulation and the insertion of the resulting measurements into the graph.
To this end, five distinctly colored markers are affixed along the continuum robot backbone, as shown in Fig.~\ref{fig:left-view}.
Because each color corresponds to a predefined arclength, stereo matching immediately resolves point correspondences, as seen in Fig.~\ref{fig:stereo-match}, directly assigning the triangulated 3D points to the appropriate camera point factors in the graph.

A primary limitation of this visual pipeline is occlusion or field-of-view exiting.
While optimal workspace coverage would require additional cameras (e.g., mounted on the distal section of the second link), in-depth sensor placement analysis remains beyond the scope of this work.
Consequently, the simulated trajectory was specifically designed to ensure that the continuum link remains continuously visible throughout the maneuver. 

\subsection{Estimator configuration}
The factor graph in Fig.~\ref{fig:hybrid-fg} was constructed and optimized using the GTSAM library.~\cite{gtsam}
The continuum appendage's strain field was parameterized using cubic spline bases with four control points for $x$- and $y$-axis bending, yielding $\bm{q}_{\sss S} \in \mathbb{R}^8$ within the estimator.~\cite{ticozzi2026}
Consistently with the number of markers, $N+1=5$ pose variables were assigned to the distal appendage. 

Table~\ref{tab:covar-val} details the numerical covariance values applied in Eq.~\eqref{eq:nlls}.
The camera point factors employed an isotropic noise model with a standard deviation of $\sigma_{\mathcal{C}_L}=5$~mm, derived from the median triangulation error of the colored backbone markers.
Finally, factors representing exact kinematic constraints ($\phi_{\sss \mathcal{J}_j}^R$, $\phi_{\sss \mathcal{Y}/\mathcal{Z}}^{\mathrm{off}}$ and $\phi_i^M$) were assigned tight covariances, while the prior covariance $\bm{\Sigma}_p$ was configured consistently with previous work.~\cite{ticozzi2026}

\begin{table}
\centering
\caption{Covariance values for the HSMS factors.}
\label{tab:covar-val}
\begin{tabular}{ll}
\toprule
Covariance & Value \\
\midrule
$\bm{\Sigma}_{\mathcal{B}}$                  & $3\cdot\mathrm{blkdiag}(\sigma^2_{\mathcal{B},\mathrm{rot}}\bm{I}_3,\, \sigma^2_{\mathcal{B},\mathrm{pos}}\bm{I}_3)$ \\
$\bm{\Sigma}_p$                              & $9\cdot10^6\bm{I}_{8}$ \\
$\bm{\Sigma}_{\mathcal{J}}^R$              & $1\cdot10^{-8}\bm{I}_6$ \\
$\Sigma_q$                                   & $3\cdot \sigma_q^2$ \\
$\bm{\Sigma}_{\mathrm{off}}$ & $1\cdot10^{-8}\bm{I}_6$ \\
$\bm{\Sigma}_M$                            & $1\cdot10^{-10}\bm{I}_6$ \\
$\bm{\Sigma}_{\mathcal{C}_L}$              & $2.5\cdot10^{-5}\bm{I}_3$ \\
\bottomrule
\end{tabular}
\end{table}

The HSMS configuration was estimated at $30$ Hz along the simulated trajectory, yielding 1{,}801 discrete timesteps.
At the initial timestep, the Levenberg-Marquardt (LM) solver was seeded with the reference configuration $\bm{q}_0= \bm{0} \in \mathbb{R}^{11}$, while subsequent LM optimizations were warm-started using the solution retrieved from the preceding timestep.

\begin{table*}[t]
\centering
\caption{Error statistics of the estimated frame poses with respect to ground truth, expressed in the world frame. The columns \textit{mean} and \textit{max} report the trajectory mean and maximum of the error magnitude $\|\bm{e}(t_k)\|$, with the mean position error additionally expressed as a percentage of the characteristic length of the system, $d = 3.905$~m, taken as the Cartesian distance between the base frame $\mathcal{B}$ and the tip frame $\mathcal{E}$ in the home configuration.
The columns \textit{emp} and \textit{pred} report the empirical and estimator-predicted standard deviations, RMS-combined across axes as $\sqrt{\mathrm{tr}\,\bm{\Sigma}}$; \textit{ratio} is their quotient.}
\label{tab:error-stats}
\begin{tabular}{lcccccccccccc}
\toprule
& \multicolumn{6}{c}{Position [mm]} & & \multicolumn{5}{c}{Orientation [deg]} \\
\cmidrule{2-7} \cmidrule{9-13}
Frame & mean & \% $d$ & max & emp & pred & ratio & & mean & max & emp & pred & ratio \\
\midrule
$\mathcal{B}$   & 16.13 & 0.41\% &  41.39 &  17.49 &  30.00 & 0.583 & &  1.60 &  4.92 & 1.73 &  3.01 & 0.577 \\
$\mathcal{R}_1$ & 37.03 & 0.95\% & 133.48 &  41.14 &  70.95 & 0.580 & &  1.60 &  4.92 & 1.73 &  3.01 & 0.576 \\
$\mathcal{R}_2$ & 66.52 & 1.70\% & 261.70 &  74.43 & 128.62 & 0.579 & &  1.60 &  4.92 & 1.73 &  3.01 & 0.577 \\
$\mathcal{F}$   & 80.39 & 2.06\% & 307.83 &  90.22 & 156.13 & 0.578 & &  1.60 &  4.92 & 1.73 &  3.01 & 0.577 \\
$\mathcal{M}$   & 87.04 & 2.23\% & 323.17 &  97.66 & 169.20 & 0.577 & &  1.76 &  6.13 & 1.92 &  3.64 & 0.527 \\
$\mathcal{E}$   & 90.03 & 2.31\% & 327.35 & 100.72 & 174.60 & 0.577 & &  4.20 & 21.88 & 4.74 & 11.84 & 0.401 \\
\bottomrule
\end{tabular}
\end{table*}

\subsection{Results}
\begin{figure}
\centering
\begin{subfigure}[b]{\linewidth}
    \centering
    \includegraphics[width=\linewidth]{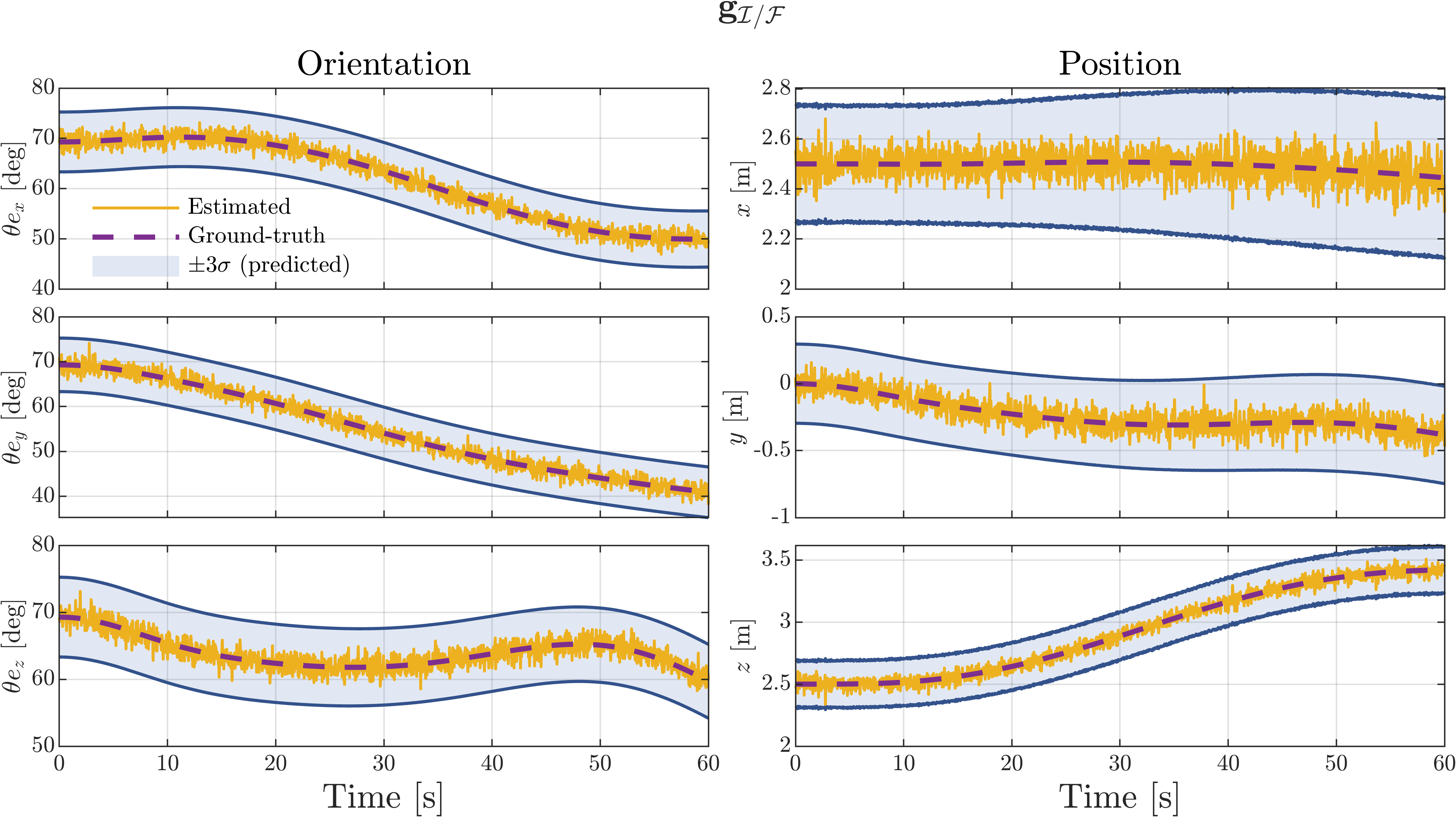}
    \caption{Frame $\mathcal{F}$.}
    \label{fig:g_I_F}
\end{subfigure}

\begin{subfigure}[b]{\linewidth}
    \centering
    \includegraphics[width=\linewidth]{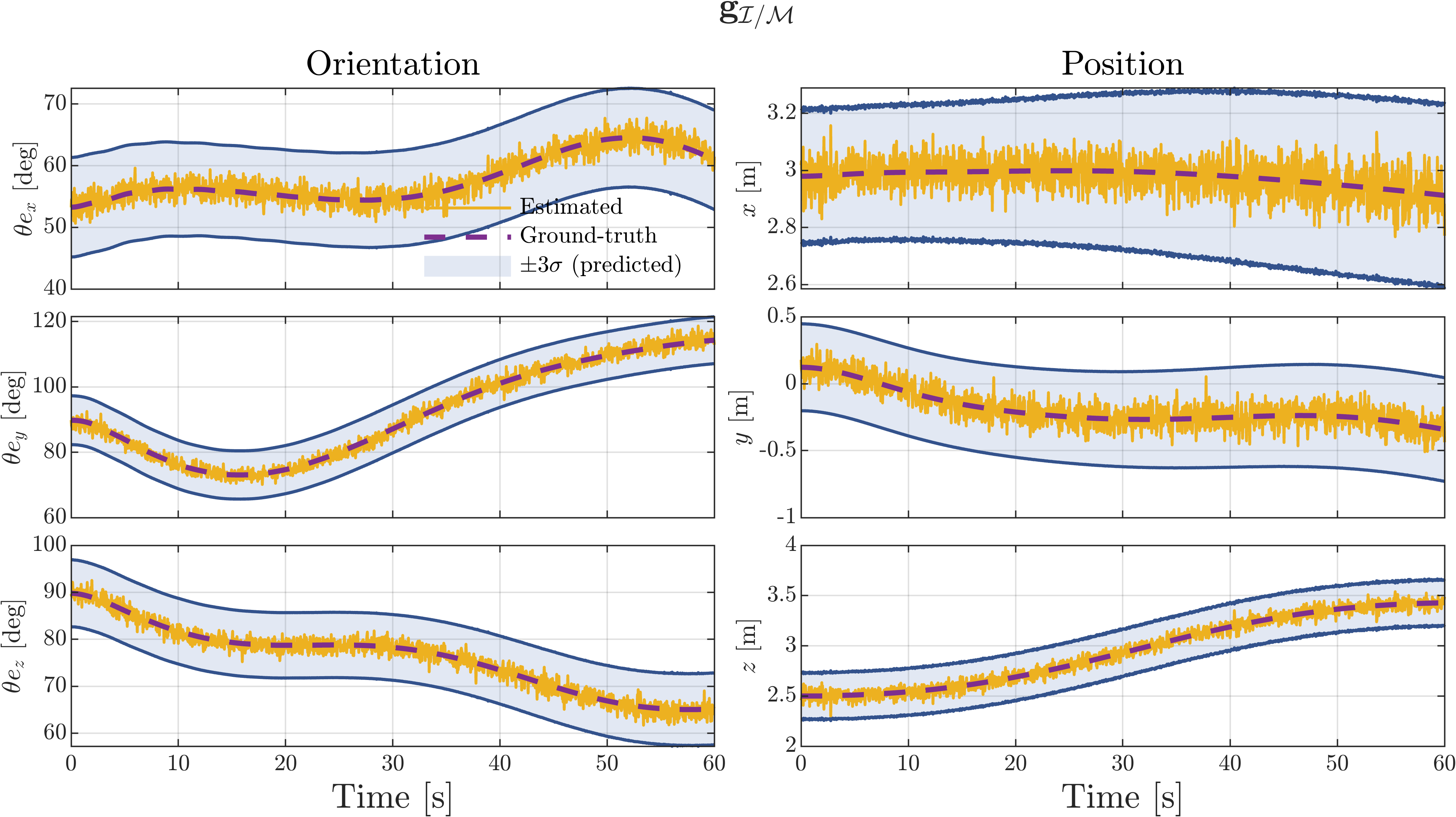}
    \caption{Frame $\mathcal{M}$.}
    \label{fig:g_I_M}
\end{subfigure}

\begin{subfigure}[b]{\linewidth}
    \centering
    \includegraphics[width=\linewidth]{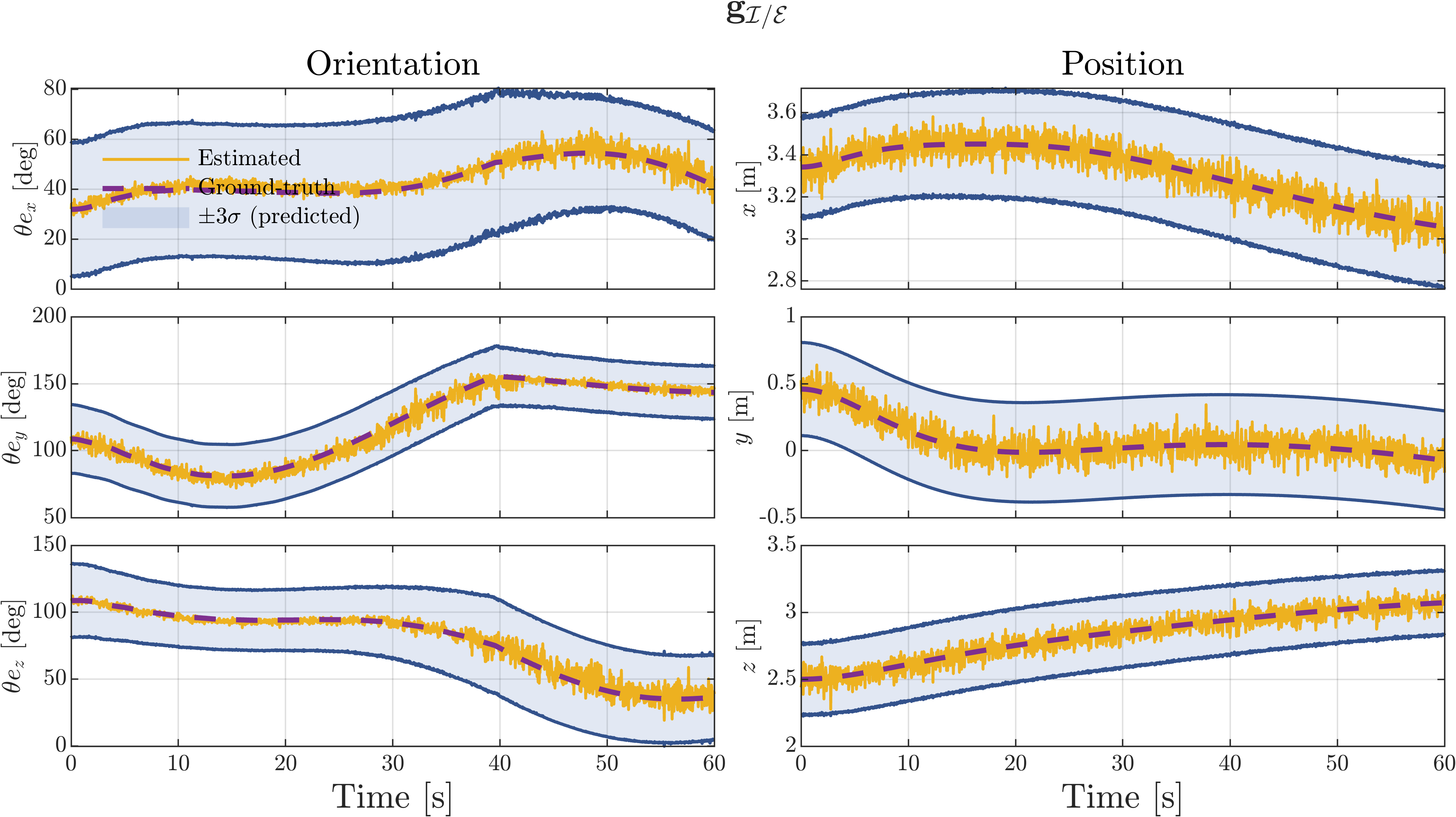}
    \caption{End-effector frame $\mathcal{E}$.}
    \label{fig:g_I_E}
\end{subfigure}
\caption{Ground-truth versus estimated pose of frames $\mathcal{F}$, $\mathcal{M}$, and $\mathcal{E}$ w.r.t. the global frame $\mathcal{I}$, with estimator-predicted $\pm3\sigma$ intervals.}
\label{fig:g_I_all}
\end{figure}

The estimation results are reported in Table~\ref{tab:error-stats} in the form of error statistics, accompanied by the ground-truth and estimated pose evolutions in Figs.~\ref{fig:g_I_F},~\ref{fig:g_I_M} and~\ref{fig:g_I_E}.

The largest position error is associated with the tip frame $\mathcal{E}$, with a trajectory mean of $90.03$~mm.
Referred to the characteristic length of the system $d = 3.905$~m, taken as the distance between $\mathcal{B}$ and $\mathcal{E}$ in the home configuration, this amounts to $2.3\%$ of $d$.
The error decreases monotonically towards the base, consistent with the accumulation of uncertainty along the kinematic chain.
The orientation error statistics are, instead, almost identical from $\mathcal{B}$ to $\mathcal{F}$, as expected for a near-deterministic rigid chain.
The encoder and rigid-link factors introduce no appreciable attitude uncertainty, so the base attitude error is propagated unchanged along the rigid links.

The ratio between the empirical and predicted standard deviations, reported in the \textit{ratio} columns of Table~\ref{tab:error-stats}, shows the consistency of the uncertainty returned by the estimator.
For the position estimates, the ratio remains close to $1/\sqrt{3} = 0.577$ at every frame, matching the inflation factor applied to the variances of $\phi_{\mathcal{B}}$ and $\phi_j^q$, thus confirming an accurate calibration of the predicted position uncertainty.
The same holds for the orientation of the rigid-body frames, confirming that the attitude estimate along the rigid part of the system is governed by the base measurement factor.
\begin{figure}
\centering
\includegraphics[width=\linewidth]{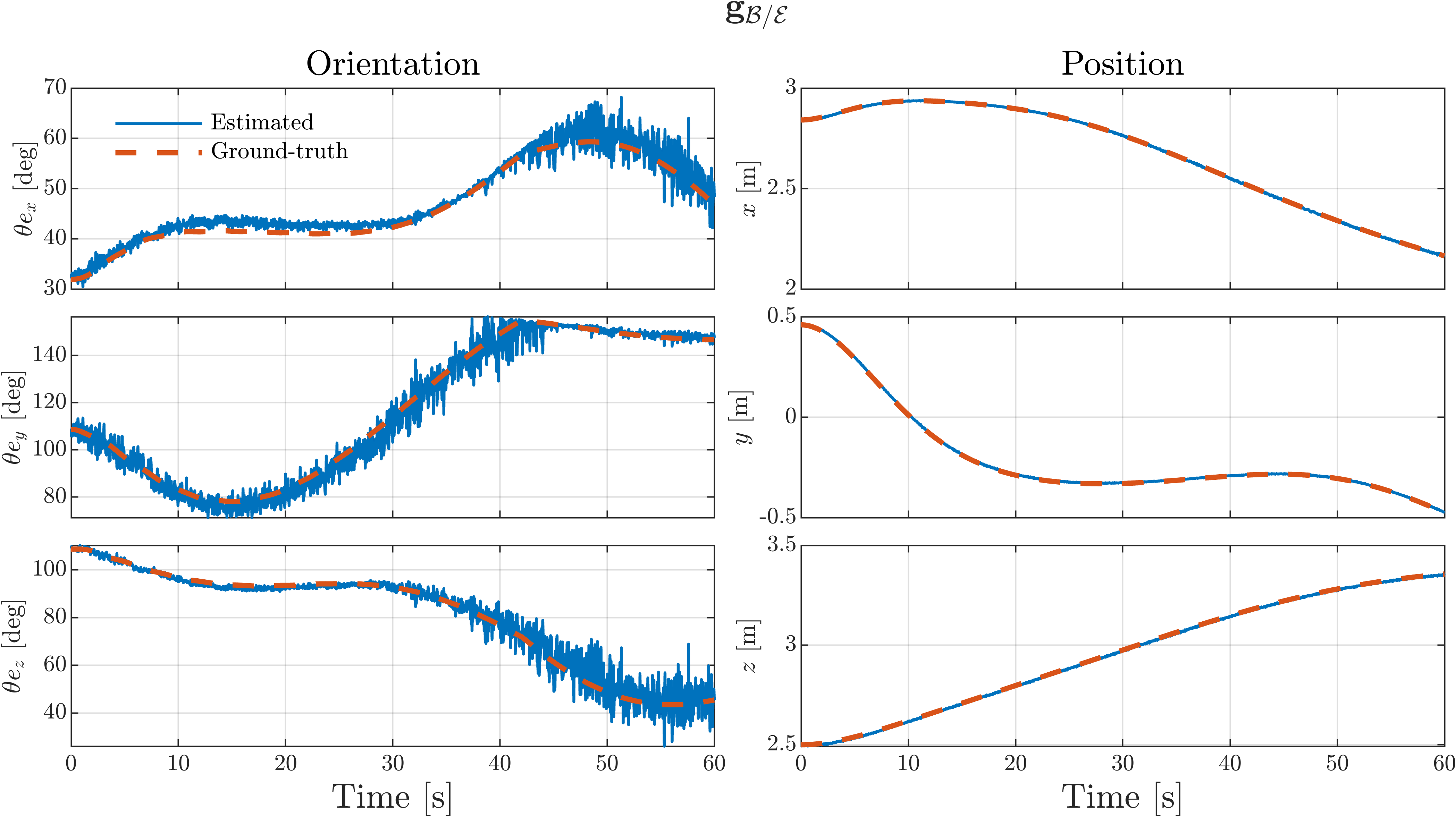}
\caption{Ground-truth versus estimated pose of the end-effector frame $\mathcal{E}$ w.r.t. the base frame $\mathcal{B}$.}
\label{fig:g_B_E}
\end{figure}
Further evidence of this fact is provided by the relative base-to-tip pose estimate $\bm{g}_{\sss \mathcal{B}/\mathcal{E}} = \bm{g}_{\sss \mathcal{I}/\mathcal{B}}^{-1} \bm{g}_{\sss \mathcal{I}/\mathcal{E}}$, displayed in Fig.~\ref{fig:g_B_E}.
Because the base pose cancels exactly in this composition, $\bm{g}_{\sss \mathcal{B}/\mathcal{E}}$ is constrained directly by the stereo measurements, independent of the base estimate.
The resulting relative position estimate is accurate to the millimeter, compared with the centimeter-level accuracy attained in the inertial frame.
The accuracy limits in the reconstruction of the absolute position of the continuum link are therefore dictated by the relative navigation solution rather than by the visual frontend.

A different behavior is observed for the orientation estimates of frames $\mathcal{M}$, the local backbone frame at $s=L/2$, and $\mathcal{E}$, characterized by mean errors of $1.76^\circ$ and $4.20^\circ$, respectively, and by increasingly conservative uncertainty bands, as shown by the last column of Table~\ref{tab:error-stats} and by Figs.~\ref{fig:g_I_M} and~\ref{fig:g_I_E}.
The $1/\sqrt{3}$ reference does not apply to these two entries, since the base prior accounts for only a minor share of their predicted attitude variance.
Such performance degradation is attributed to the visual setup used, which carries no direct information about the orientation of the continuum appendage.

Finally, the computational performance of the proposed algorithm, run on a 14-core Apple M4 Pro chip, is reported in Table~\ref{tab:comp-perf}, with an average solution time per timestep of the order of $1$~ms.

\section{Conclusion}
This paper presented a novel approach to reconstructing the configuration of a hybrid soft-rigid kinematic chain from discrete, noisy measurements, specialized to a hybrid soft-rigid space manipulator system.
The methodology is predicated on the use of factor graphs and leverages the Geometric Variable Strain framework, whose strain coefficients are themselves estimation variables.
Validation along a $60$-sec closed-loop maneuver showed a mean end-effector position error of $2.3\%$ of the characteristic length of the system, reducing to the millimeter level relative to the base frame, and an average solution time of $1.23$~ms.
Beyond the point estimate, the optimization returns a marginal covariance for every variable, whose predicted position uncertainty was found to be accurately calibrated across the entire chain.
Future work should address the absence of direct attitude information on the continuum appendage, investigate optimal sensor configurations, and test the proposed formulation on representative hardware.

\begin{table}[t]
\centering
\caption{Computational performance data.}
\label{tab:comp-perf}
\begin{tabular}{ll}
\toprule
Timesteps & 1{,}801 \\
\midrule
Avg solve time & 1.230 ms/timestep  \\
\midrule
Avg LM iterations  & 7.271 \\
\midrule
Max solve time  & 4.462 ms \\
\midrule
Max LM iterations  & 24 \\
\midrule
Total solve time & 2.216 s \\
\bottomrule
\end{tabular}
\end{table}

\section{Statements and declarations}

\subsection{Ethical considerations}
Not applicable.

\subsection{Consent to participate}
Not applicable.

\subsection{Consent for publication}
Not applicable.

\subsection{Declaration of conflicting interest}
The author(s) declared no potential conflicts of interest with respect to the research, authorship, and/or publication of this article.

\section{Funding}
The author(s) disclosed receipt of the following financial support for the research, authorship, and/or publication of this article: This work was supported by the AFOSR SURI award FA9550-23-1-0723 and by NSF FRR award 2101250.

\section{Supplemental material}
Supplemental material for this article is available online.



\begin{thebibliography}{10}
\providecommand{\url}[1]{\texttt{#1}}
\providecommand{\urlprefix}{URL }
\expandafter\ifx\csname urlstyle\endcsname\relax
  \providecommand{\doi}[1]{DOI:\discretionary{}{}{}#1}\else
  \providecommand{\doi}{DOI:\discretionary{}{}{}\begingroup
  \urlstyle{rm}\Url}\fi
\providecommand{\eprint}[2][]{\url{#2}}

\bibitem{wef2024}
Forum WE.
\newblock Space: The \$1.8 trillion opportunity for global economic growth,
  2024.
\newblock \urlprefix\url{https://www3.weforum.org/docs/WEF_Space_2024.pdf}.
\newblock Accessed: 2026-07-27.

\bibitem{tsiotras2023}
Tsiotras P, King-Smith M and Ticozzi L.
\newblock Spacecraft-mounted robotics.
\newblock \emph{Annual Review of Control, Robotics, and Autonomous Systems}
  2023; 6: 335--362.

\bibitem{dubowsky1993}
Dubowsky S and Papadopoulos E.
\newblock The kinematics, dynamics, and control of free-flying and
  free-floating space robotic systems.
\newblock \emph{IEEE Transactions on Robotics and Automation} 1993; 9(5):
  531--543.

\bibitem{valverde2018}
Valverde A and Tsiotras P.
\newblock Dual quaternion framework for modeling of spacecraft-mounted
  multibody robotic systems.
\newblock \emph{Frontiers in Robotics and AI} 2018; Volume 5 - 2018.
\newblock \doi{10.3389/frobt.2018.00128}.

\bibitem{senda1995}
Senda K, Murotsu Y, Nagaoka H et~al.
\newblock Attitude control for free-flying space robot with {CMG} (control
  moment gyroscopes).
\newblock In \emph{Guidance, Navigation, and Control Conference}. AIAA, pp.
  1494--1502.

\bibitem{giordano2016}
Giordano AM, Garofalo G, De~Stefano M et~al.
\newblock Dynamics and control of a free-floating space robot in presence of
  nonzero linear and angular momenta.
\newblock In \emph{2016 IEEE 55th Conference on Decision and Control (CDC)}.
  pp. 7527--7534.
\newblock \doi{10.1109/CDC.2016.7799432}.

\bibitem{antonello2019}
Antonello A, Valverde A and Tsiotras P.
\newblock Dynamics and control of spacecraft manipulators with thrusters and
  momentum exchange devices.
\newblock \emph{Journal of Guidance, Control, and Dynamics} 2019; 42(1):
  15--29.
\newblock \doi{10.2514/1.G003601}.

\bibitem{doggett2014}
Doggett WR, Dorsey JT, Jones TC et~al.
\newblock Development of a tendon-actuated lightweight in-space manipulator
  (talisman).
\newblock In \emph{42nd Aerospace Mechanisms Symposium}. NASA Goddard Space
  Flight Center, MD.

\bibitem{dorsey2015}
Dorsey J, Doggett WR, Jones TC et~al.
\newblock Application of a novel long-reach manipulator concept to asteroid
  redirect missions.
\newblock In \emph{2nd AIAA Spacecraft Structures Conference}. Kissimmee, FL.

\bibitem{li2019}
Li K, Zhang Y and Hu Q.
\newblock Dynamic modelling and control of a tendon-actuated lightweight space
  manipulator.
\newblock \emph{Aerospace Science and Technology} 2019; 84: 1150--1163.
\newblock \doi{https://doi.org/10.1016/j.ast.2018.11.018}.

\bibitem{kernot2022}
Kernot JE and Ulrich S.
\newblock Adaptive control of a tendon-driven manipulator for capturing
  non-cooperative space targets.
\newblock \emph{Journal of Spacecraft and Rockets} 2022; 59(1): 111--128.
\newblock \doi{10.2514/1.A34881}.

\bibitem{yang2022}
Yang J, Peng H, Zhang J et~al.
\newblock Dynamic modeling and beating phenomenon analysis of space robots with
  continuum manipulators.
\newblock \emph{Chinese Journal of Aeronautics} 2022; 35(9): 226--241.
\newblock \doi{10.1016/j.cja.2021.08.007}.

\bibitem{yang2022jgcd}
Yang J, Peng H, Zhou W et~al.
\newblock Integrated control of continuum-manipulator space robots with
  actuator saturation and disturbances.
\newblock \emph{Journal of Guidance, Control, and Dynamics} 2022; 45(12):
  2379--2388.
\newblock \doi{10.2514/1.G006692}.

\bibitem{yang2024}
Yang J, Peng H, Wu S et~al.
\newblock Differential-algebraic equation-based model predictive control for
  trajectory tracking of spacecraft-mounted continuum manipulators.
\newblock \emph{Aerospace Science and Technology} 2024; 155: 109615.
\newblock \doi{10.1016/j.ast.2024.109615}.

\bibitem{mathew2024}
Mathew AT, Feliu-Talegon D, Alkayas AY et~al.
\newblock Reduced order modeling of hybrid soft-rigid robots using global,
  local, and state-dependent strain parameterization.
\newblock \emph{The International Journal of Robotics Research} 2025; 44(1):
  129--154.
\newblock \doi{10.1177/02783649241262333}.

\bibitem{ticozzi2025}
Ticozzi L, Vela PA and Tsiotras P.
\newblock Modeling and control of a hybrid soft–rigid space manipulator
  system.
\newblock \emph{Journal of Guidance, Control, and Dynamics} 2025; 48(10):
  2162--2179.
\newblock \doi{10.2514/1.G009231}.

\bibitem{lilge2022}
Lilge S, Barfoot TD and Burgner-Kahrs J.
\newblock Continuum robot state estimation using {Gaussian} process regression
  on {SE(3)}.
\newblock \emph{The International Journal of Robotics Research} 2022;
  41(13-14): 1099--1120.
\newblock \doi{10.1177/02783649221128843}.

\bibitem{ticozzi2026}
Ticozzi L, Vela PA and Tsiotras P.
\newblock Factor graph-based shape estimation for continuum robots via {M}agnus
  expansion, 2026.
\newblock \urlprefix\url{https://arxiv.org/abs/2604.15619}.
\newblock \eprint{2604.15619}.

\bibitem{renda2020}
Renda F, Armanini C, Lebastard V et~al.
\newblock A geometric variable-strain approach for static modeling of soft
  manipulators with tendon and fluidic actuation.
\newblock \emph{IEEE Robotics and Automation Letters} 2020; 5(3): 4006--4013.
\newblock \doi{10.1109/LRA.2020.2985620}.

\bibitem{magnus1954}
Magnus W.
\newblock On the exponential solution of differential equations for a linear
  operator.
\newblock \emph{Communications on Pure and Applied Mathematics} 1954; 7(4):
  649--673.
\newblock \doi{10.1002/cpa.3160070404}.

\bibitem{blanes1998}
Blanes S, Casas F, Oteo JA et~al.
\newblock Magnus and {F}er expansions for matrix differential equations: the
  convergence problem.
\newblock \emph{Journal of Physics A: Mathematical and General} 1998; 31(1):
  259--268.
\newblock \doi{10.1088/0305-4470/31/1/023}.

\bibitem{dellaert2017}
Dellaert F and Kaess M.
\newblock Factor graphs for robot perception.
\newblock \emph{Found Trends Robot} 2017; 6(1–2): 1–139.
\newblock \doi{10.1561/2300000043}.

\bibitem{mathew2023}
Mathew AT, Hmida IB, Armanini C et~al.
\newblock Sorosim: A {MATLAB} toolbox for hybrid rigid–soft robots based on
  the geometric variable-strain approach.
\newblock \emph{IEEE Robotics \& Automation Magazine} 2023; 30(3): 106--122.
\newblock \doi{10.1109/MRA.2022.3202488}.

\bibitem{opromolla2017}
Opromolla R, Fasano G, Rufino G et~al.
\newblock A review of cooperative and uncooperative spacecraft pose
  determination techniques for close-proximity operations.
\newblock \emph{Progress in Aerospace Sciences} 2017; 93: 53--72.

\bibitem{gtsam}
Dellaert F and {GTSAM Contributors}.
\newblock borglab/gtsam, 2022.
\newblock \urlprefix\url{https://github.com/borglab/gtsam}.

\end{thebibliography}
\end{document}